\documentclass[dvipsnames]{article}

\usepackage[nonatbib, final]{neurips_2023}

\usepackage[T1]{fontenc}    
\usepackage{rotating}
\usepackage[para]{footmisc}
\usepackage{tablefootnote}
\usepackage{hyperref}  
\usepackage{amssymb}
\usepackage{url}            
\usepackage{booktabs}       
\usepackage{amsfonts}       
\usepackage{nicefrac}       
\usepackage{microtype}      
\usepackage{xcolor}         
\usepackage{bbding}         
\usepackage{soul}
\usepackage{cleveref}
\usepackage{graphicx}
\usepackage{subcaption}
\usepackage{multirow}
\usepackage[para, flushleft]{threeparttable}
\usepackage{dirtytalk}
\usepackage[inline]{enumitem}
\usepackage{stackengine}
\hypersetup{linktoc=all}

\graphicspath{{Figures/}{../Figures/}}

\usepackage{scalerel,xparse}

\NewDocumentCommand\mrc{}{
    \includegraphics[scale=0.06]{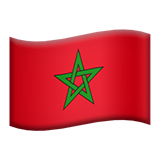}
}

\NewDocumentCommand\tns{}{
    \includegraphics[scale=0.06]{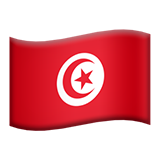}
}

\NewDocumentCommand\alg{}{
    \includegraphics[scale=0.06]{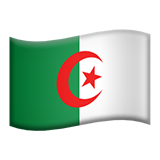}
}

\newcommand\homograph{%
\includegraphics[scale=0.28]{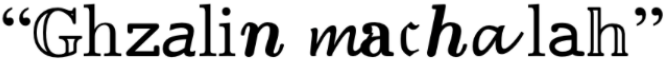}}

\title{Homograph Attacks on Maghreb Sentiment Analyzers}

\author{Fatima Zahra Qachfar \\
fqachfar@uh.edu \\
University of Houston
\And Rakesh M. Verma\\
rmverma2@central.uh.edu\\
University of Houston}

\begin{document}

\maketitle

\begin{abstract}
    We examine the impact of homograph attacks on the Sentiment Analysis (SA) task of different Arabic dialects from the Maghreb North-African countries. Homograph attacks result in a 65.3\% decrease in transformer classification from an F1-score of 0.95 to 0.33 when data is written in \say{\textit{Arabizi}}. The goal of this study is to highlight LLMs weaknesses' and to prioritize ethical and responsible Machine Learning. 

\end{abstract}

\section{Introduction}



In recent years, large language models (LLMs) have received considerable attention due to their adaptability and success in various NLP tasks, including text classification and sentiment analysis. Sentiment analysis is a technique used by companies to optimize their marketing strategies by gaining a deeper understanding of their customers' emotions towards their products and services. However, this peak of interest has also made LLMs susceptible to malicious attacks such as homograph attacks. 
 

Homograph attacks are character-based perturbation attacks that rely on visually similar characters to deceive individuals into believing they are accessing authentic content. Despite their identical appearance, these characters have distinct Unicode identifiers that set them apart. An \textit{Arabizi} example from the MDMD \cite{Younoussi_Azroumahli_2022} test set can be perturbed from 
\say{Ghzalin macha lah} to \homograph.




These attacks pose significant challenges for LLMs that accept raw unfiltered input text, making them vulnerable to Unicode manipulation. For example, we give ChatGPT 3.5 the same instruction with and without homograph perturbation. ChatGPT was not able to generate the same meaningful response asking the user to rephrase their prompt.



In this study, we examine the impact of this attack on Maghreb sentiment analyzers of North Africa which encompasses four countries: \textit{Morocco}, \textit{Tunisia}, \textit{Algeria}, and \textit{Libya}. 
Maghreb dialects are unstructured low-resource languages that present different challenges than Modern Standard Arabic.
The following difficulties in Maghreb dialect processing directly correlate with the homograph attack:

\textbf{Code-Switching:} refers to the practice of switching between languages within a text. This phenomenon can occur because of cultural and historical factors such as colonization. 


\textbf{Arabizi Scripts:} refers to the use of the Latin script to represent Arabic language text, instead of Arabic alphabet. This writing style is often used in informal context in SMS, and social media posts, especially among younger generations. 




To summarize, we make the following contributions:
\begin{enumerate*}[label= \textbf{\roman*)}]
    \item Showing the impact of character based attack on North-African languages, and
    \item Highlighting the importance of ethical and responsible ML. 
\end{enumerate*}


\section{Methodology and Evaluation} 

\textbf{Methodology :} 
We preform a homograph perturbation attack on the testing sets, while keeping the training set intact. Each example in the test set is perturbed with a  percentage of 90\% of Latin characters with Unicode homographs.


\textbf{Datasets :} We choose challenging \textit{SA} datasets which includes Arabizi, and Arabic written dialects with code-switching. As illustrated in \Cref{tab:datasets}, these datasets contain short sentences. 
When combining Tunisian dialect datasets (TUNIZI \cite{Chayma2020}, and TSAC \cite{medhaffar-etal-2017-sentiment}), we create the same label2id mapping to ensure consistent labels.
We split the datasets into 80\% training 10\% validation, and 10\% testing.

\begin{table}[!ht]
\small
\centering
\caption{Maghreb Arabic Dialect Datasets for Sentiment Analysis}
\scalebox{0.88}{
\begin{threeparttable}
\begin{tabular}{lccccc}

\toprule
\textbf{Dataset} & \textbf{Total Size} & \textbf{N\_Classes} & \textbf{Distribution} & \textbf{Avg Length} & \textbf{Type-Token Ratio}  \tnote{\textbf{$\ast$}}
\\ \midrule
MDMD \cite{Younoussi_Azroumahli_2022} \mrc
 & $ 42,000 $  & 3 (pos/neut/neg) & Extremely Imbalanced \tnote{\textbf{$\ddagger$}} & $\sim 5$ Tokens & $0.31$
\\ 

MDMD[\textbf{DAL}] \cite{Younoussi_Azroumahli_2022} \mrc  & $ 18,184 $  & 3 (pos/neut/neg) & Extremely Imbalanced \tnote{\textbf{$\ddagger$}} & $\sim 5$ Tokens & $ 0.37 $
\\ 

TUNIZI  \cite{Chayma2020} \tns & $ 3,000 $ & 2 (pos/neg) & Balanced & $\sim 8$ Tokens & $ 0.44 $
\\

TSAC \cite{medhaffar-etal-2017-sentiment} \tns & $ 11, 871 $ & 2 (pos/neg) & Slightly Imbalanced & $\sim 9$ Tokens & $ 0.29 $
\\

AlgD \cite{mazari2022sentiment} \alg & $ 11,760  $ & 3 (pos/neut/neg) & Slightly Imbalanced & $\sim 19$ Tokens & $0.22$

\\ \bottomrule

\end{tabular}
\begin{tablenotes}
\centering
\item[\textbf{$\ddagger$}] \textit{Extremely Imbalanced Class is Neutral Labeled Sentiment}
\item[\textbf{$\ast$}] \textit{TTR : Lexical Diversity Dataset Measure After Emoji Removal}
\end{tablenotes}

\end{threeparttable}
}

\label{tab:datasets}
\end{table}


\textbf{Evaluation :} We have chosen pre-trained language models that support North-African dialects for both Arabic and Latin characters. We fine-tune these models on the datasets for sentiment analysis with a learning rate of $2e^{-5}$ during three training epochs. We apply the homograph attack on the testing set to report the impact of this attacks on the classification performance 

To further show the impact of such attacks, we conduct one experiment on only Arabizi data and another on mix of Arabizi, Arabic written dialects with code-switching.

\begin{table}[!ht]
\small
\centering
\caption{Impact of Homograph Attack on Text Classification On North-African Dialect Languages. The evaluation metrics include F1-score (F1), Precision (P), Recall (R), and Accuracy (Acc).}
\scalebox{0.85}{
\begin{threeparttable}
\begin{tabular}{llcccc}

\toprule
\textbf{Model} & \textbf{Dataset} & \textbf{F1 (BA/AA)} & \textbf{P (BA/AA)} & \textbf{R  (BA/AA)} & \textbf{Acc (BA/AA)} 
\\ \midrule



DarijaBERT-mix \tablefootnote{\scriptsize \url{https://huggingface.co/SI2M-Lab/DarijaBERT-mix}} & MDMD \cite{Younoussi_Azroumahli_2022} \mrc & ($0.77$/$0.58$) ($\textbf{\textcolor{red}{24.7\%}} \triangledown$) & ($0.78$/$0.70$)  & ($0.77$/$0.56$)  & ($0.77$/$0.64$) 
\\ 

DarijaBERT-arabizi \tablefootnote{\scriptsize \url{https://huggingface.co/SI2M-Lab/DarijaBERT-arabizi}} & MDMD[\textbf{DAL}] \cite{Younoussi_Azroumahli_2022} \mrc & ($0.75$/$0.30$)
 ($\textbf{\textcolor{red}{60.0\%}} \triangledown$) & ($0.76$/$0.30$) &  ($0.75$/$0.40$) &  ($0.77$/$0.40$)
\\ 

TunBERT \tablefootnote{\scriptsize  \url{https://huggingface.co/ziedsb19/tunbert_zied} \hspace{20pt}} & TUNIZI  \cite{Chayma2020} + TSAC \cite{medhaffar-etal-2017-sentiment} \tns  & ($0.78$/$0.51$) ($\textbf{\textcolor{red}{34.6\%}} \triangledown$) & ($0.78$/$0.57$) & ($0.78$/$0.55$) & ($0.78$/$0.54$) \\

TunBERT & TUNIZI  \cite{Chayma2020} \tns &  ($0.95$/$0.33$) ($\textbf{\textcolor{red}{65.3\%}} \triangledown$) & ($0.95$/$0.25$) & ($0.95$/$0.50$)& ($0.95$/$0.49$)\\


DziriBERT \tablefootnote{\scriptsize  \url{https://huggingface.co/alger-ia/dziribert}} & AlgD \cite{mazari2022sentiment} \alg & ($0.86$/$0.57$)  ($\textbf{\textcolor{red}{33.7\%}} \triangledown$)  & ($0.86$/$0.66$) & ($0.86$/$0.61$) & ($0.86$/$0.59$)
\\ \bottomrule



\end{tabular}

\begin{tablenotes}
\centering
\item[] \textit{\textbf{BA} : Before Attack.}
\item[] \textit{\textbf{AA} : After Attack.}
\item[] \textit{\textbf{DAL} : Dialect Arabic in Latin.}
\end{tablenotes}

\end{threeparttable}
}

\label{tab:results}
\end{table}

In \Cref{tab:results}, we observe that the highest performance decrease of 60\%, and 65.3\% respectively on models fine-tuned on {MDMD[DAL]} \cite{Younoussi_Azroumahli_2022}, and {TUNIZI} \cite{Chayma2020} belong to dataset with \textit{only Arabizi} writing scripts which means that all the samples were affected by this attack. The lowest performance decrease of 24.7\% although less than the other decreases, is still noteworthy and considerable. The size of the dataset and distribution of Latin and Arabic written samples can affect this percentage.

\section{Conclusions and Future Works}

When developing responsible machine learning systems, it is crucial to consider the potential harm of malicious manipulations like homograph attacks. In this paper, we show the impact of such attacks on low-resource Maghreb dialect languages to promote the importance of ethical and trustworthy ML.


By taking into account similar attacks, we will be able to create resilient models that are less prone to tampering and can face adversarial attempts. In future works, we will focus on developing effective defense mechanisms to safeguard LLMs against homograph attacks. To address this challenge, our first attempt consists of designing a protection input layer for LLMs by reverse engineering the perturbation approach where we first identify Unicode characters and correct them.




\section*{Acknowledgments} Research partly supported by NSF grants 2210198 and 2244279, and ARO grants W911NF-20-1-0254 and W911NF-23-1-0191. Verma is the founder of Everest Cyber Security and Analytics, Inc.

\bibliographystyle{plain}
{\footnotesize \bibliography{references}}

\end{document}